# A COMPARATIVE STUDY OF HYPERPARAMETER TUNING METHODS

SUBHASIS DASGUPTA & JAYDIP SEN

## Introduction

The world is changing fast towards automation and artificial intelligence (AI). At the heart of AI rests different machine learning models. It is a known fact that each machine learning model has its bias-variance trade-off (Briscoe & Feldman, 2011; Doroudi, 2020). This trade-off is important to make sure that the model is more generalized. Model generalization is important because both underfit and overfit models would render an AI system less effective. Machine learning models recognize patterns by estimating the model parameters which in turn get affected by the choice of the hyperparameters. It is to be understood that not all model possesses hyperparameters (basic naïve Bayes algorithm). However, it is true that under different training situations, different models get inclined to more bias error or more variance error. In this context, let us understand the concept of bias error and variance error in the area of regression analysis. Let the actual value of the target be $Y$ for the input $X$. Also, let us assume that the model under consideration was trained with m different training samples without changing the set of hyperparameters associated with the model. Let the set of parameters of the model be defined by $\theta$. If $F(X|\theta)$ denotes the model output for the input $X$, we can write

$$F(X|\theta) = Y$$

Now, since the model was trained m times with *m* different training samples, we can write,

$$F_i(X|\theta_i) = Y_i, \forall\, i \in \{1,2,3,\dots m\}$$

`

Let $\hat{Y}$ denotes the average of all the individual model's predictions. Thus, $\hat{Y} = \frac{1}{m}\sum_{i=1}^{m} Y_i$. If we take the difference between $Y$ and $\hat{Y}$, $(Y - \hat{Y})$ is defined as a *bias error*. Again, since there are m predictions of $Y$, we can calculate the variance of the predictions by calculating $\sigma^2 = \frac{1}{m}\sum_{i=1}^{m}(Y_i - \hat{Y})^2$. This $\sigma^2$ constitutes the variance error. It is possible to prove that mean square error is essentially the sum of the bias² and variance error.

$$MSE = \frac{1}{m}\sum_{i=1}^{m}(Y - Y_i)^2$$

$$MSE = \frac{1}{m}\sum_{i=1}^{m}(Y^2 - 2YY_i + Y_i^2)$$

$$MSE = \frac{1}{m}\left(\sum_{i=1}^{m} Y^2 - \sum_{i=1}^{m} 2YY_i + \sum_{i=1}^{m} Y_i^2\right)$$

$$MSE = \frac{1}{m}\left(mY^2 - 2Y\sum_{i=1}^{m} Y_i + \sum_{i=1}^{m} Y_i^2\right)$$

$$MSE = \frac{1}{m}\left(mY^2 - 2Y(m\hat{Y}) + \sum_{i=1}^{m} Y_i^2\right)$$

$$MSE = \frac{1}{m}\left(mY^2 - 2Y(m\hat{Y}) + m(\hat{Y})^2 + \sum_{i=1}^{m} Y_i^2 - m(\hat{Y})^2\right)$$

$$MSE = \frac{1}{m}(mY^2 - 2Y(m\hat{Y}) + m(\hat{Y})^2) + \frac{1}{m}\left(\sum_{i=1}^{m} Y_i^2 - m(\hat{Y})^2\right)$$

$$MSE = (Y^2 - 2Y(\hat{Y}) + (\hat{Y})^2) + \frac{1}{m}\left(\sum_{i=1}^{m} Y_i^2 - m(\hat{Y})^2\right)$$

`

$$MSE = (Y - \hat{Y})^2 + \frac{1}{m}\left(\sum_{i=1}^{m} Y_i^2 - m(\hat{Y})^2\right)$$

$$MSE = bias^2 + variance$$

Thus, for a given MSE if bias is reduced, the variance error will start increasing and vice versa. This is where the bias-variance trade-off becomes important. The graph of bias-variance trade-off in Figure 8.1 shows that at the intersection of the bias error and the variance error the model has the lowest generalization error.

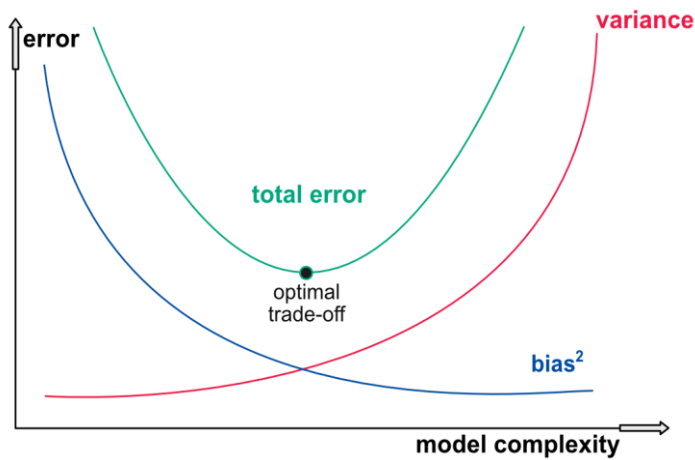

**Figure 8.1.** Bias Variance Trade-off

[adapted from https://www.mlfactor.com/images/var_bias_trade2.png]

Finding this optimal trade-off point is not easy. The time complexity starts increasing very quickly as the number of hyperparameters to be optimized starts increasing. This leads to a difficult situation for data analysts to come up with the best model. Statistical methods are mostly parametric, and hence, they may possess high bias for certain datasets where the pattern to be recognized is quite non-linear. Similarly, algorithm-based non-parametric machine learning models may become too complex to memorize the training data to produce a highly overfit model. There are several methods suggested in the literature to tackle this problem but in this work, only a handful of the most frequently used methods are going to be discussed at length.

## Related work

As mentioned in the previous section, different researchers, over time, have suggested several methods to tackle the problem of hyperparameter tuning. The most commonly used algorithms are grid search and random search (Bergstra & Bengio, 2012). However, these search processes are, rather, not intelligent enough to use the information from previous searches. But, it does not mean that they are always inferior to other methods. Grid search, for example, is always going to give the best combination of hyperparameters if sufficient resources can be allocated. Sufficient resources mean the availability of lots of parallel computing processors to evaluate the combination of hyperparameters independently of other combinations. If the researcher/analyst is well versed with the hyperparameters and their implications on the model, he/she can choose the sets carefully to get the best results. Particularly, if the number of hyperparameters to be tuned is no more than three and sufficient resources are available with parallel computing, grid search can produce a really good set of hyperparameters for the model under consideration. Random search, on the other hand, is computationally cheaper than grid search. If more time is consumed to search the hyperparameters, as per the concept of Monte Carlo, the probability of selecting the best set of hyperparameters is also increased. This characteristic is absent in the case of the grid search process. Random search is probabilistic and hence getting the best result is not guaranteed. However, in practice, this search process is found to perform better than grid search. Even though random search tends to perform better than grid search, it is still computationally intensive. That is why more sophisticated search algorithms are proposed by different researchers. Some of the algorithms are described below.

`

*Genetic Algorithm*

One of the many such algorithms is the genetic optimization process. Genetic algorithms work based on evolutionary concepts. The evolutionary process iterates through a few critical steps such as:

1. Embedding the hyperparameters in a digital chromosome
2. Creating the initial population of chromosomes
3. Calculating fitness scores of each chromosome
4. Selecting chromosomes for subsequent operations, i.e. crossover and mutation
5. Use crossover to create a new chromosome out of the existing two chromosomes (exploiting the existing information for local search)
6. Use mutation to alter the chromosome randomly (exploring the search space for global search)
7. Continue the process from step 3 until the maximum iteration limit is reached or if the improvement is not beyond a threshold value for a fixed number of iterations

For an elaborate explanation of the process, the readers can refer to Mallawaarachchhi (Mallawaarachchi, 2017) and (Katoch et al., 2021). This algorithm is meta-heuristic and it has been used in many different domains. If a genetic algorithm is allowed to run for a reasonably long time, it tries to reach the global optimal solution or near global optimal solution even if the objective function is highly non-convex. Moreover, this algorithm is capable of optimizing multiple objectives simultaneously. That is why, this method is used in many different domains such as the Travelling Sales Man problem (Larranaga et al., 1999), image enhancement and segmentation (Paulinas & Ušinskas, 2007), construction planning and scheduling (Nusen et al., 2021), container allocation in cloud (Guerrero et al., 2018) and many more. Because of its special characteristics, the same algorithm is used in the domain of machine learning where it is used for optimizing the hyperparameters of different models (Gorgolis et al., 2019; Han et al., 2020; Lessmann et al., 2005; Tayebi & El Kafhali, 2021; Vincent & Jidesh, 2023).

*Tree-structured Parzen Estimation*

`

Tree-structured parzen estimation (TPE) is a very popular hyperparameter tuning process. Bayesian optimization process focuses on estimating the posterior probability based on the information available on likelihood and prior probability. TPE proposes a different utility function that acts as a surrogate in this Bayesian optimization process. Let us assume that the score of the objective function is $y$ for a given set of hyperparameters $x$. As per Baye's rule,

$$p(y|x) = \frac{p(x|y)p(y)}{p(x)} \quad (1)$$

Expected improvement (EI) in the context of TPE (Bergstra et al., 2011) can be evaluated based on (2).

$$EI_{y^*}(x) = \int_{-\infty}^{y^*} (y^* - y)p(y|x)dy \quad (2)$$

where $y^*$ is current optima. Instead of evaluating $p(y|x)$ directly using Sequential Model-Based Optimization (SMBO) (Hutter et al., 2011), TPE defines the likelihood as a tree structure as given in (3).

$$p(x|y) = \begin{cases} l(x), & if\ y < y^* \\ g(x), & if\ y \geq y^* \end{cases} \quad (3)$$

Here, $l(x)$ stands for the density function of $x$ such that the corresponding outcome $y$ is below some threshold $y^*$ and $g(x)$ stands for the density function when $y$ is more than the threshold value. TPE doesn't bother about any specific model for $p(y)$.

If $\gamma = p(y < y^*)$, the probability of occurrence of $x$ can be defined as $p(x) = \gamma l(x) + (1 - \gamma)g(x)$. Hence, expected improvement (EI) becomes,

$$EI = \int_{-\infty}^{y^*} (y^* - y)p(y|x)dy$$

$$= \int_{-\infty}^{y^*} (y^* - y)\frac{p(x|y)p(y)}{p(x)}dy$$

`

$$= \int_{-\infty}^{y^*} (y^* - y) \frac{l(x)p(y)}{p(x)} dy$$

$$= \frac{1}{p(x)} \left[ y^* l(x) \int_{-\infty}^{y^*} p(y) dy - l(x) \int_{-\infty}^{y^*} y p(y) dy \right]$$

$$= \frac{1}{p(x)} \left[ y^* l(x) \gamma - l(x) \int_{-\infty}^{y^*} y p(y) dy \right]$$

$$= \frac{y^* l(x) \gamma - l(x) \int_{-\infty}^{y^*} y p(y) dy}{\gamma l(x) + (1 - \gamma) g(x)}$$

$$= \frac{y^* \gamma - \int_{-\infty}^{y^*} y p(y) dy}{\gamma + \frac{g(x)}{l(x)} (1 - \gamma)}$$

Thus, it can be seen that $EI \propto \left[ \gamma + \frac{g(x)}{l(x)} (1 - \gamma) \right]^{-1}$. The numerator is independent of both $l(x)$ and $g(x)$. In other words, the EI will attain a maximum value if the ratio $\frac{l(x)}{g(x)}$ is maximized. This would mean that the samples of $x$ are to be picked more from the $l(x)$ distribution than the $g(x)$ distribution. This maximization is much cheaper from the computational point of view than evaluating the score of the objective function. Hence, instead of optimizing the objective function, the surrogate function (EI) is maximized and the hyperparameter set is supplied to evaluate the actual score $y$. With the increase in y, the distributions $l(x)$ and $g(x)$ are determined more accurately and $\gamma$ is decided as some percentile score of all available $y$ scores. Hence, with the increase in the iteration, TPE starts producing a more optimum set of hyperparameters. There are several research works where TPE has been used (Khoei et al., 2021; Liang et al., 2022; Ozaki et al., 2020; Shen et al., 2022; Zhao & Li, 2018) and researchers have done comparative analysis also (Putatunda & Rama, 2018).

`

# Methodology

In this work, 4 datasets are considered from the UCI Machine Learning repository (Asuncion & Newman, 2007) where 2 of them are meant for the task of regression analysis, and the remaining 2 are for classification analysis. These datasets have a reasonably large number of data points and a relatively lesser number of variables. Hence, dealing with the variables is easier for these datasets and this is important for the current study as the focus is lying on the relative performances of hyperparameter tuning processes rather than analyzing the data for business insights. The brief dataset descriptions are given in Table 8.1 below. Three of these datasets are clean in the sense that they do not have any missing values whereas one dataset has some missing values.

**TABLE 8.1.** BASIC DATASET DESCRIPTION

| Sl No | Task | Name of the dataset | # of Instances | # of features | Has missing data? |
|---|---|---|---|---|---|
| 1 | Regression | Gas turbine CO and NOx emission data | 36733 | 11 | No |
| 2 | Regression | Steel industry energy consumption | 35040 | 11 | No |
| 3 | Classification | Adult | 48842 | 14 | Yes |
| 4 | Classification | Dry Bean Dataset | 13611 | 17 | No |

The only focus maintained in this work is to analyze the relative performance of different machine learning models while their hyperparameters are tuned using the methods mentioned above. For this study, the models considered are:

- For regression analysis:

`

1. Regularized linear regression (for regression analysis)
2. Ada-boost regression
3. RandomForest regression
4. Gradient Boosting Machine regression
5. XgBoost regression
6. Light GBM regression

- For classification analysis
    1. Regularized logistic regression
    2. Ada-boost classification
    3. RandomForest classification
    4. Gradient Boosting Machine classification
    5. XgBoost classification
    6. Light GBM classification

Linear regression and logistic regression are considered to create linear models for comparison purposes. Other models have larger sets of hyperparameters, and a better comparison can be found if a larger search space is considered while finding the optimal set of hyperparameters. The experiments are done using Python-based packages. For grid and random search, scikit-learn (Kramer & Kramer, 2016) is used. For genetic, Bayesian, and TPE optimization, the optuna package (Akiba et al., 2019) is used. For simulated annealing-based hyperparameter tuning, the GitHub repository of SantoshHari (Hari, 2018) is used. Root Mean Square Error (RMSE) is considered as the metric for regression whereas Area Under the ROC Curve (AUC) is considered as the metric for the classification tasks. 3-fold cross-validation is used to evaluate the performances of the models with different sets of hyperparameters. The hyperparameters of different models with their corresponding levels are mentioned in the appendix. All the experiments are done on the Google Colab platform without any GPU support.

## Analysis of the Results

Before proceeding with the model-building exercises, it is important to perform exploratory data analysis (EDA). In the following portion, basic EDAs are explained. Deep-down EDA is not

`

performed as the objective of this study is to compare the hyperparameter tuning processes. Hence, basic EDAs and basic preprocessing of data are performed.

*Basic EDA of Steel industry energy consumption dataset*

The dataset on the steel industry deals with power generation in kWh along with other variables. There are situations when the load requirements are less whereas there are situations when the load requirements increase significantly. The objective associated with this dataset is to predict the power requirements when other variables are given. The dataset is split into two parts with a 70:30 ratio to create training and test datasets. This step is necessary so that the performance of all the models can be verified for a common test set. The model hyperparameters are to be trained based on the 3-fold cross-validation on the training dataset. The variables associated with the dataset are given in Table 8.2.

The dataset contains categorical variables, and they are converted to dummy variables using a one-hot-encoding process. For any model to work properly, it is important to check if the data distributions of the training set and the test set are similar or not. Because, otherwise, the model will tend to commit mistakes while predicting the values for the test data. The distribution of the data in the training and the test sets is shown in Figure 8.3. This dataset has a time stamp for each data point and hence while splitting the data into training and test sets, data points are not shuffled randomly. The splitting ratio is kept at 70:30 for training and test datasets.

As per the distribution plot, the data distribution for the training and test datasets are similar and hence, if a model is trained on the training set, the same should work equally well on the test set. However, for certain variables, the distributions differed, at least visually, such as LagCP. If this variable turns out to be critical in predicting the load usage, then the model will suffer greater losses while predicting the usage for the test dataset. That is why, understanding variable importance turns out to be critical in such situations.

`

The scatter plot shown in Figure 8.2 shows that the usage and some of the predictor variables show some linear correlations between them but not all the variables have linear correlation with the target variable. This also suggests that nonlinear models might produce better results than linear models. The linear regression model is very good in understanding the impact of a predictor variable on the target variable. But linear regression model comes with a lot of assumptions and hence the associated biases.

**TABLE 8.2.** VARIABLE DESCRIPTION OF STEEL INDUSTRY DATA

| Variable | Abbreviation | Type | Measurement |
| --- | --- | --- | --- |
| Date | Date | Continuous | date |
| Industry energy consumption | KhW | Continuous | kWh |
| Lagging current reactive power | LagRP | Continuous | kVarh |
| Leading current reactive power | LeadRP | Continuous | kVarh |
| $tCO_2(CO_2)$ | $CO_2$ | Continuous | Ppm |
| Lagging current power factor | | Continuous | % |
| Leading current power factor | LagCP | Continuous | % |
| Number of seconds from midnight | LeadCP | Continuous | S |
| Week status | Wstat | Categorical | Weekend (0) or a Weekday (1) |
| Day of the week | Day | Categorical | Sunday, Monday, ..., Saturday |
| Load type | Load_type | Categorical | Light Load, Medium Load, Maximum Load |

However, if interactions are considered, then a linear model might turn out to be better than the nonlinear models. But that study is out of the scope of the current study and hence no efforts are put to understand interaction effects on model predictions.

`

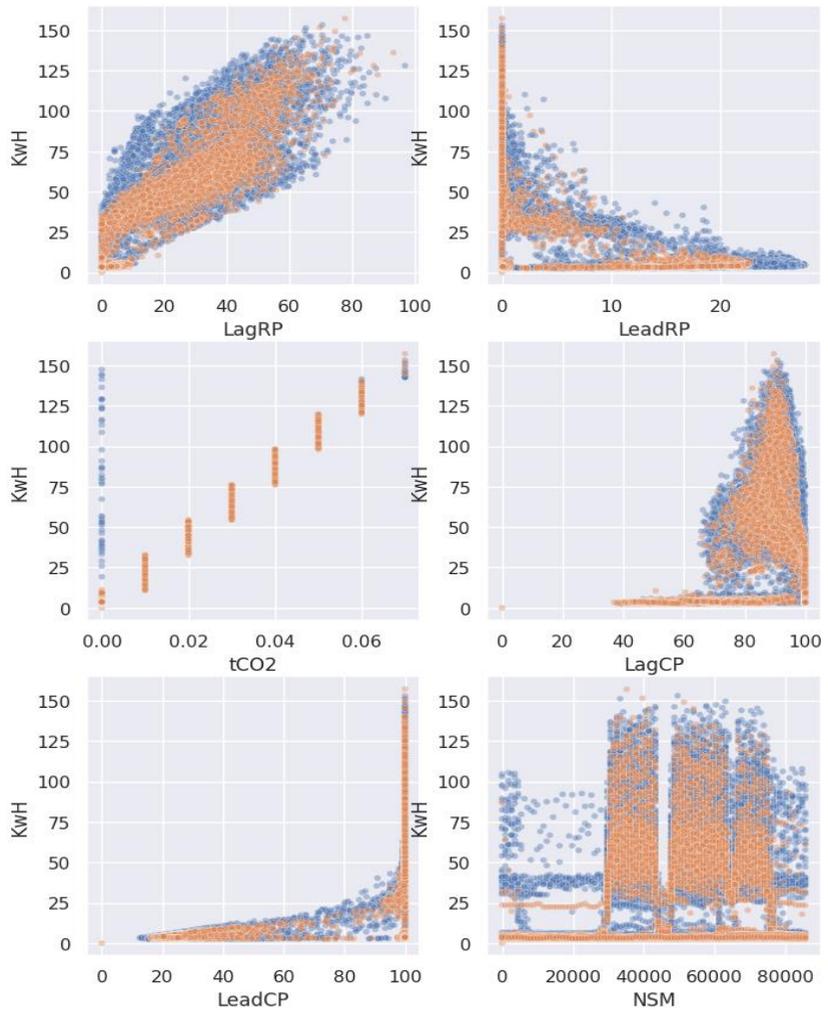

**Figure 8.2**. Scatter plot between kWh and other numerical variables in the training and the test dataset

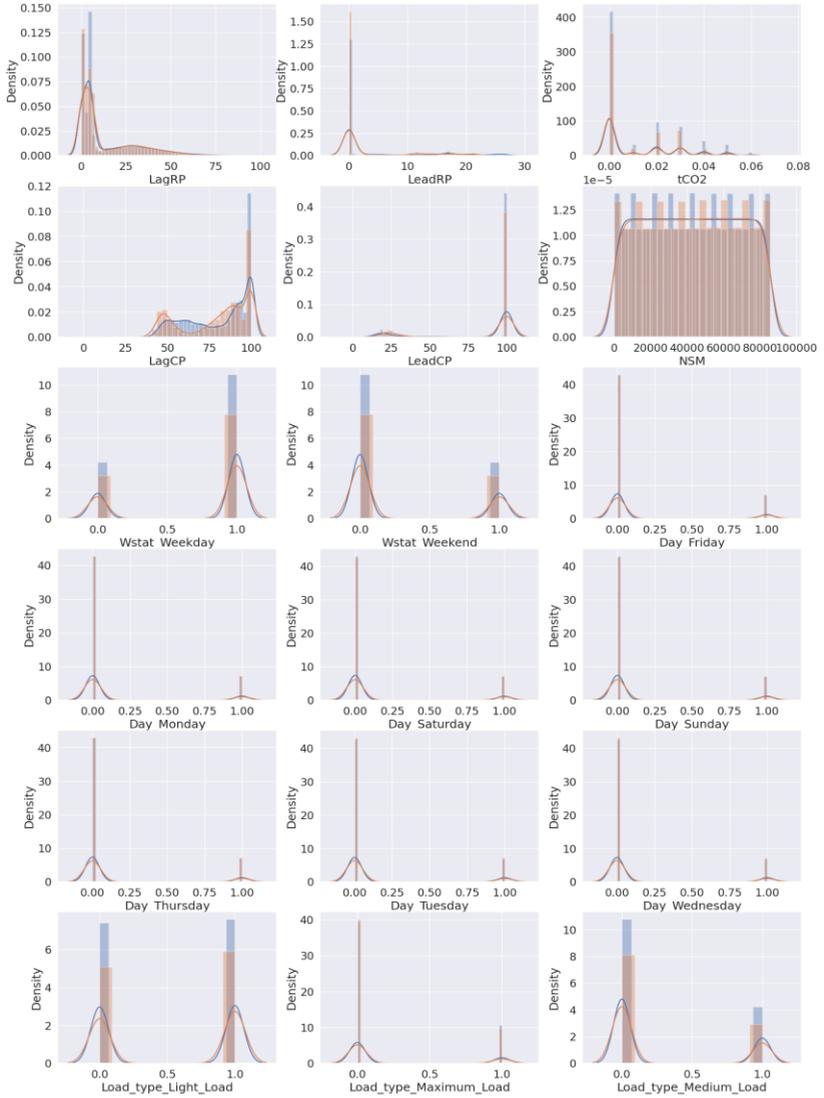

**Figure 8.3**. Distribution of data in training and test set for steel industry dataset. The blue colour represents the training dataset, and the brown colour indicates test dataset.

*Basic EDA of gas turbine dataset*

The dataset on the gas turbine power generator is related to the emission of carbon monoxide (CO) and nitrogen oxide ($NO_x$) through the combustion of fuel in the gas turbine plant. These two gases are toxic and their release to the atmosphere can lead to health problems. Hence, it is important to monitor the release of these gases and it is also helpful to predict how much gases will be released during the operation based on other operative parameters. This dataset does not have any time stamp and hence random splitting of the dataset is more meaningful. This dataset is also split based on a 70:30 ratio for training and testing. The variable description is given in the Table 8.3 below.

**TABLE 8.3.** VARIABLE DESCRIPTION OF GAS TURBINE DATA

| Variable | Abbreviation | Type | Measurement |
|---|---|---|---|
| Ambient temperature | AT | Continuous | C |
| Ambient pressure | AP | Continuous | mbar |
| Ambient humidity | AH | Continuous | % |
| Air filter difference pressure | AFDP | Continuous | mbar |
| Gas turbine exhaust pressure | GTEP | Continuous | mbar |
| Turbine inlet temperature | TIT | Continuous | C |
| Turbine after temperature | TAT | Continuous | C |
| Compressor discharge pressure | CDP | Continuous | mbar |
| Turbine energy yield | TEY | Continuous | MWH |
| Carbon monoxide | CO | Continuous | $mg/m^3$ |
| Nitrogen oxide | $NO_x$ | Continuous | $mg/m^3$ |

This dataset has two target variables and hence two separate models can be created to predict the output of CO and $NO_x$. However, in this study, the prediction of CO is considered as this is more dangerous compared to $NO_x$. The distribution of predictor variables in the training set and test set is shown in Figure 8.4. Incidentally, all the predictor variables are continuous, and hence no encoding is

`

required for these variables. Also, as per Figure 8.4, most of the variables are multi-modal. The variables, with multi-modal distribution show 3 modal distributions. Thus, the dataset may have 3 clusters. But, without running a cluster analysis, nothing can be said concretely. Similarly, a scatterplot can be seen to understand how the target variable is related to the predictor variables. As per Figure 8.5, CO looks quite closely related to different variables such as AFDP, GTEP, TIT, TAT, CDP, and TEY. However, the relationships are not linear for all the variables. Still, a linear model might perform reasonably well in predicting the CO level based on the other process parameters. The linear relationship between two variables can be evaluated by looking at the Pearson Correlation coefficient and a correlation matrix is a good matrix to look at while trying to understand the relationships among multiple numerical variables. The correlation matrix is shown in Table 8.4. This correlation matrix shows the correlation of variables in the training set (upper triangular matrix) and the validation set (lower triangular matrix). The last variable is CO (the target variable) and it can be seen that this variable has a good correlation with other variables. Having a good amount of linear correlation with the target variable is a desirable property as far as linear models are concerned. But a closer look at the table reveals that some of the predictor variables are also correlated with each other. This is what we understand as the multicollinearity of variables. These multicollinearities create a lot of issues within the model's predictive power. For example, in case of linear regression, model's parameters get inflated and also unreliable. Not only that, the individual impact of independent variables on the dependent variable becomes quite vague due to the presence of multicollinearities. That is why some specific measures are required to be taken to reduce the impact of multicollinearity while model building. Using regularizing hyperparameters is one of the most used methods. However, there are some algorithms which are inherently somewhat robust to the presence of multicollinearity. Tree based models are not affected greatly by this one issue. That is why, tree based models are more popular than affine function based models where removal of variables during model building exercise if not advised.

`

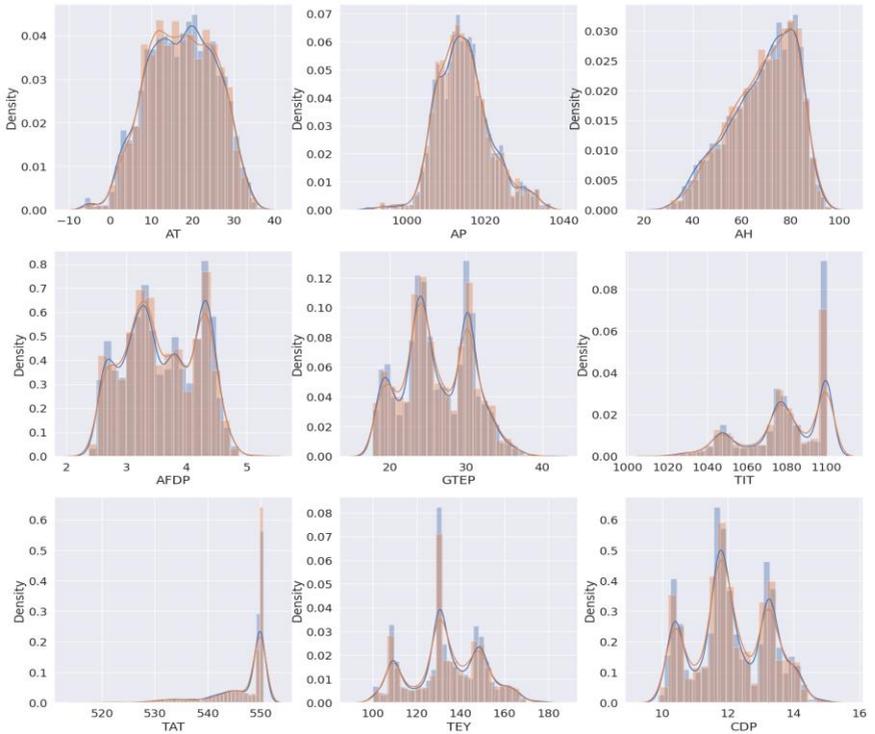

**Figure 8.4.** Distribution of predictor variables of the gas turbine data in both training and the test set

An interesting aspect of this dataset is that the predictor variables are mostly multimodal. Such distribution can be modeled as mixtures of Gaussian distributions and linear models might not properly extract the pattern. Such distribution suggests the possibility of using nonlinear modeling techniques.

`

**TABLE 8.4.** CORRELATION MATRIX FOR THE GAS TURBINE DATASET. THE UPPER TRIANGULAR MATRIX IS MEANT FOR THE TRAINING SET AND THE LOWER TRIANGULAR MATRIX IS FOR THE VALIDATION SET

|  | AT | AP | AH | AFDP | GTEP | TIT | TAT | TEY | CDP | CO |
|---|---|---|---|---|---|---|---|---|---|---|
| AT | 1.000 | -0.498 | -0.469 | 0.464 | 0.190 | 0.328 | 0.221 | 0.104 | 0.195 | -0.394 |
| AP | -0.480 | 1.000 | 0.089 | -0.086 | -0.032 | -0.073 | -0.304 | 0.063 | 0.042 | 0.193 |
| AH | -0.460 | 0.072 | 1.000 | -0.244 | -0.295 | -0.262 | 0.017 | -0.182 | -0.220 | 0.172 |
| AFDP | 0.481 | -0.114 | -0.248 | 1.000 | 0.845 | 0.915 | -0.517 | 0.885 | 0.923 | -0.651 |
| GTEP | 0.203 | -0.071 | -0.305 | 0.841 | 1.000 | 0.893 | -0.621 | 0.934 | 0.939 | -0.569 |
| TIT | 0.334 | -0.102 | -0.258 | 0.915 | 0.891 | 1.000 | -0.394 | 0.951 | 0.951 | -0.752 |
| TAT | 0.178 | -0.255 | 0.048 | -0.526 | -0.620 | -0.401 | 1.000 | -0.635 | -0.657 | 0.025 |
| TEY | 0.122 | 0.020 | -0.186 | 0.885 | 0.929 | 0.954 | -0.631 | 1.000 | 0.991 | -0.628 |
| CDP | 0.216 | 0.000 | -0.227 | 0.924 | 0.935 | 0.954 | -0.655 | 0.991 | 1.000 | -0.623 |
| CO | -0.383 | 0.220 | 0.128 | -0.617 | -0.529 | -0.705 | 0.028 | -0.590 | -0.588 | 1.000 |

## Basic EDA of Adult dataset

The adult dataset is meant for a classification analysis where the objective is to predict if a person has an income level above $50K or not. A good model predicting the income level of people is quite helpful in several business use cases. Since products are mostly developed keeping the socio-economic standard of the targeted customers, a model that can predict the economic condition of the customer, can help the marketing managers to do market segmentation more accurately for targeted campaigns with lesser costs. From this perspective, this dataset is quite interesting to analyze. Moreover, this dataset contains both numerical and categorical variables, making the model-building exercise a little bit more difficult. To add to the complexity, this dataset has missing values as well. The variable description of the dataset is given below in Table 8.5.

`

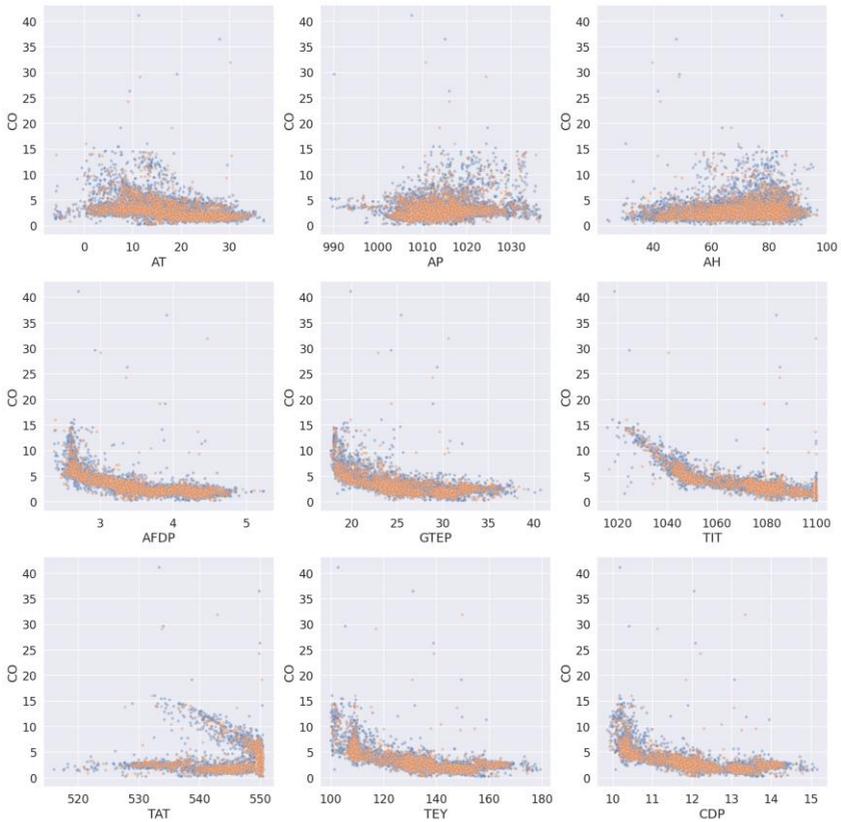

**Figure 8.5.** Scatter plot between CO and other numerical variables in the training and the test dataset of gas turbine data

**TABLE 8.5.** VARIABLE DESCRIPTION OF ADULT DATASET

| Variable Name | Demographic | Description | Missing Values |
|---|---|---|---|
| age | Age | N/A | no |
| workclass | Income | Private, Self-emp-not-inc, Self-emp-inc, Federal-gov, Local-gov. and other work classes | yes |
| fnlwgt | | | no |
| education | Education Level | Bachelors, Some-college, 11th, HS-grad, Prof-school, Assoc-acdm, Assoc-voc. and other education | no |
| education-num | Education Level | | no |
| marital-status | Other | Married-civ-spouse, Divorced, Never-married, Separated, Widowed, Married-spouse-absent, Married-AF-spouse. | no |
| occupation | Other | Tech-support, Craft-repair, Other-service, Sales, Exec-managerial, Prof-specialty, Handlers-cleaners, Machine-op-inspect and other occupations | yes |
| relationship | Other | Wife, Own-child, Husband, Not-in-family, Other-relative, Unmarried. | no |
| race | Race | White, Asian-Pac-Islander, Amer-Indian-Eskimo, Other, Black. | no |
| sex | Sex | Female, Male. | no |
| capital-gain | | | no |
| capital-loss | | | no |
| hours-per-week | | | no |
| native-country | Other | United States, Cambodia, England, Puerto Rico, Canada, Germany, Outlying-US(Guam-USVI-etc), India, Japan, Greece, South, and other countries | yes |
| income | Income | >50K, <=50K. | no |

`

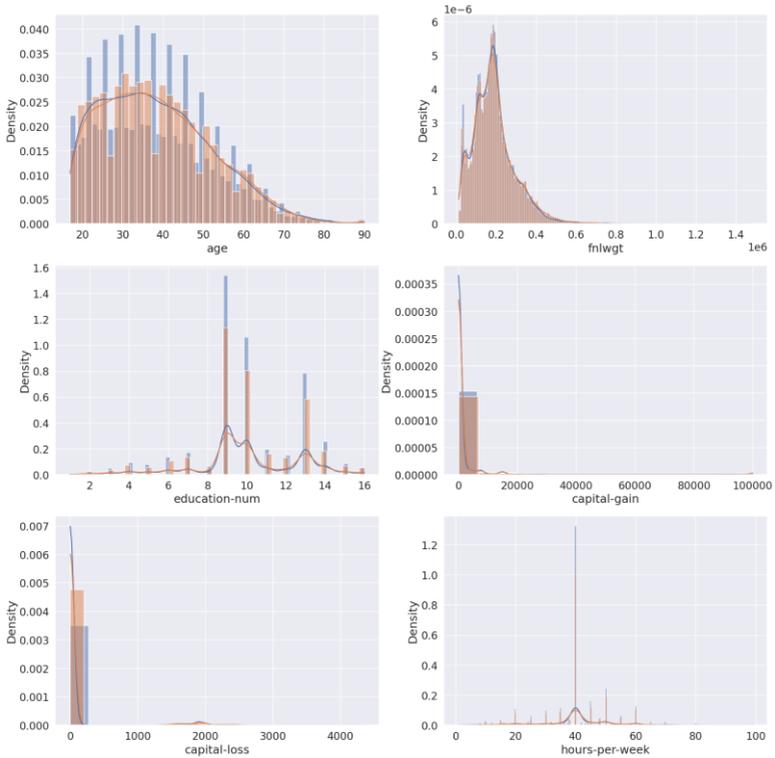

**Figure 8.6.** Distribution of numerical variables in the training and test dataset of the Adult dataset

The dataset has 5 numeric variables and remaining all the variables are categorical. Some of these categorical variables also have missing values. Like the previous datasets, this dataset is also broken into training and validation sets with 70:30 ratios. The distributions of the numerical variables are shown in Figure 8.6. It can be seen that the distributions are matching in training and the test datasets. A similar analysis is done for the categorical variables. For categorical variables, bar plots are plotted for the training and the test datasets.

The distributions of the categorical variables are shown in Figure 8.7. Here also the distributions are quite matching. Three variables contain missing values and they are imputed using a model-based iterative imputation method called MICE (Multiple Imputation by Chained Equation). A decision tree classifier is used as a base model to impute the missing values.

A decision tree is used in place of a Random Forest or any other ensemble method due to resource constraints. After data imputation, the imputed data is kept ready for the model-building process.

*Basic EDA of Dry Bean Dataset*

Computer vision has many applications in real-time analysis. The same can be applied to agricultural products as well. The dataset under consideration is meant for analyzing the seven classes of dry beans. 16 features could help do the classification. The variable descriptions are given the Table 8.6. The dataset has no missing values, and it is comprised of only numerical features. This dataset is built based on the measurements taken from the seeds. But the seeds can be analysed by simply analysing the images of the seeds also. That process leads to computer vision and computer vision is more technical in nature and is outside the scope of this chapter. If the data distribution is observed carefully in Figure 8.8, it can be observed that most of the variables are multimodal in nature. Presence of multimodal distribution gives an indication that the data are probably clustered and, maybe, linear models can perform reasonably good on the dataset for a classification task. This dataset deals with dry beans which belong to 7 different classes. Multiclass classification is also trickier in nature when the classes are imbalanced in nature. Simple accuracy score, in case of imbalanced dataset, gets biased towards the class having the highest frequency of occurrence.

`

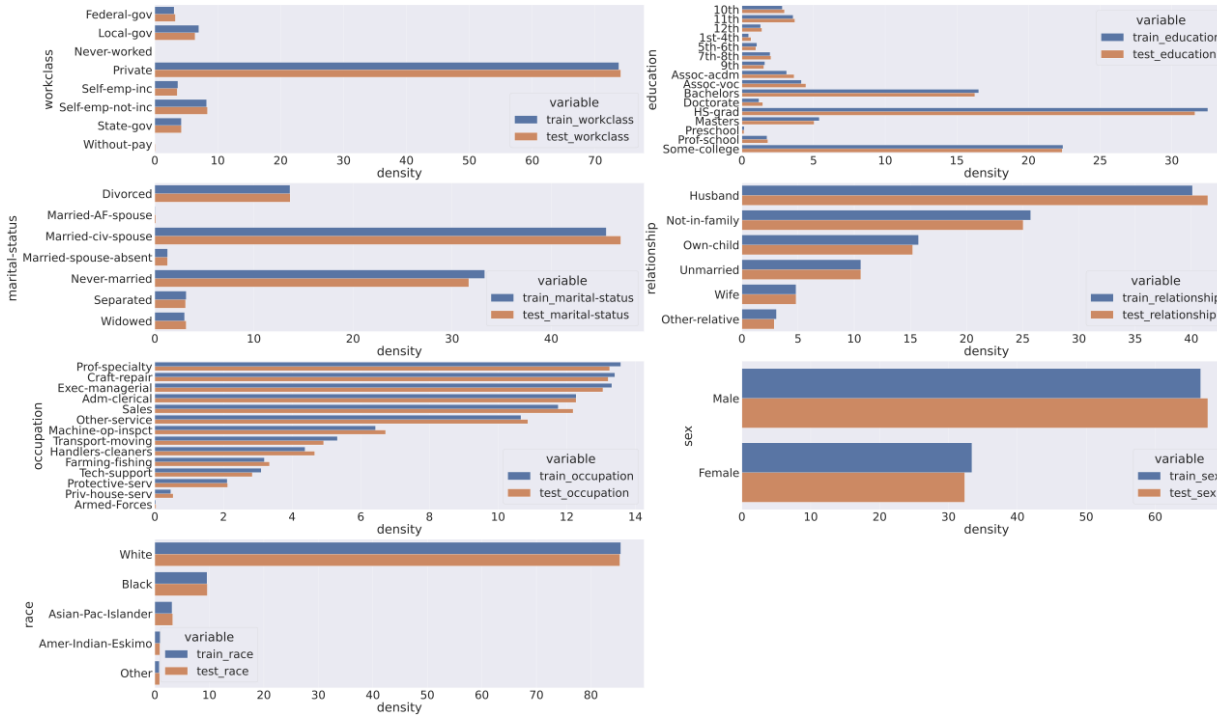

Figure 8.7: Distribution of the categorical variables of the Adult dataset in the training and the test datasets

**TABLE 8.6.** A BRIEF DESCRIPTION OF THE VARIABLES OF THE DRY BEAN DATA

| Variable Name | Type | Description |
|---|---|---|
| Area | Integer | The area of a bean zone and the number of pixels within its boundaries |
| Perimeter | Continuous | Bean circumference is defined as the length of its border. |
| MajorAxisLength | Continuous | The distance between the ends of the longest line that can be drawn from a bean |
| MinorAxisLength | Continuous | The longest line that can be drawn from the bean while standing perpendicular to the main axis |
| AspectRatio | Continuous | Defines the relationship between MajorAxisLength and MinorAxisLength |
| Eccentricity | Continuous | The eccentricity of the ellipse having the same moments as the region |
| ConvexArea | Integer | Number of pixels in the smallest convex polygon that can contain the area of a bean seed |
| EquivDiameter | Continuous | Equivalent diameter: The diameter of a circle having the same area as a bean seed area |
| Extent | Continuous | The ratio of the pixels in the bounding box to the bean area |
| Solidity | Continuous | Also known as convexity. The ratio of the pixels in the convex shell to those found in beans. |
| Roundness | Continuous | Calculated with the following formula: (4piA)/(P^2) |
| Compactness | Continuous | Measures the roundness of an object |
| ShapeFactor1 | Continuous | |
| ShapeFactor2 | Continuous | |
| ShapeFactor3 | Continuous | |
| ShapeFactor4 | Continuous | |
| Class | Categorical | (Seker, Barbunya, Bombay, Cali, Dermosan, Horoz and Sira) |

`

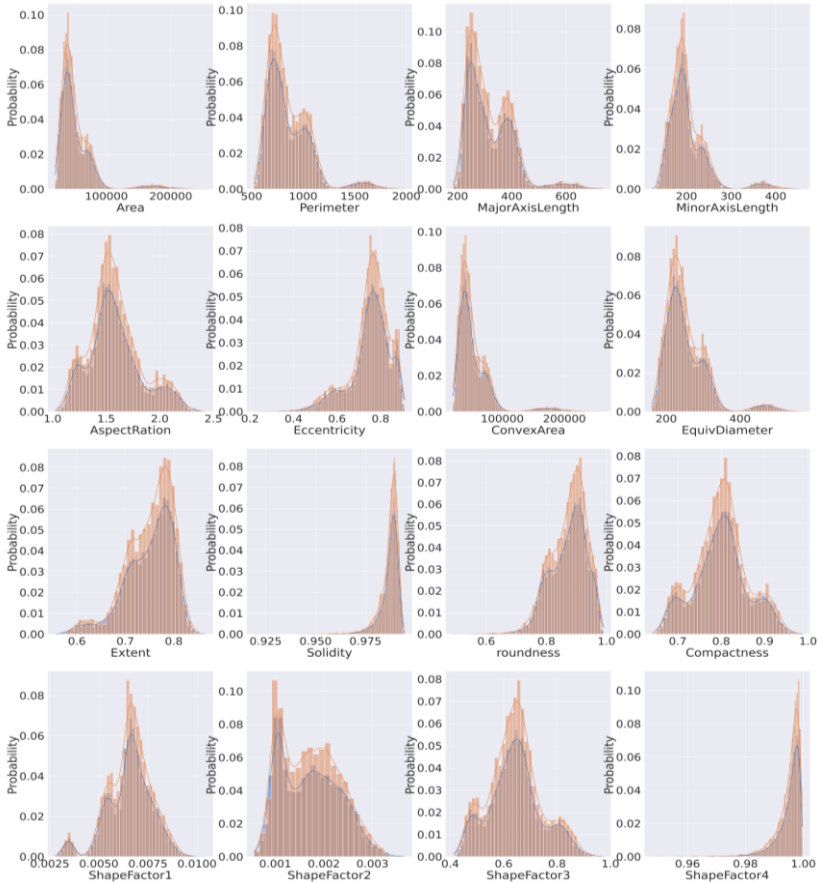

**Figure 8.8.** Distribution of data in training set and test set of the dry bean dataset

The target variable here is the prediction of the class of the 'Class' variable. There are 7 different outcomes of this variable. The dataset is split into a 70:30 ratio to create the training and test dataset. The distribution of the variables in the training and the test dataset is shown in Figure 8.8.

`

# Hyperparameter tuning

For the present study, only three methods are tested on multiple datasets. The methods are implemented in the Optuna package (Akiba et al., 2019). Optuna is a dedicated package for performing hyperparameter optimization using various methods and this package is kept updated by the contributors. Hence, for the experimentation part, this package is used. As mentioned earlier, different machine learning algorithms work with different sets of hyperparameters and hence sets of different hyperparameters are required to be supplied to the respective models so that the best set of hyperparameters can be selected. The set of hyperparameters for different models is mentioned in the table below.

It is to be noted that the values are going to be picked up from distributions and hence, theoretically, there are an infinite number of combinations of hyperparameters possible from which the best combination is to be picked for each model. This far more comprehensive search than what can be done with the Grid Search option. Hence, Grid Search is kept outside the comparison.

However, one should be aware that a grid search is a good option if the search space is small so that the number of evaluations is less.

Table 8.8 shows the performance of the algorithms of the regression analysis on the steel industry dataset. The number of iterations is kept at 100 for each method. Interestingly, the random search provided the best result with the lowest RMSE value. A similar result is also seen in the case of the gas turbine dataset as shown in Table 8.9. Random search provided the best combination of hyperparameters. However, if the outcomes of classification are considered in Table 8.10 and Table 8.11, TPE turned out to be the winner. The genetic algorithm followed TPE closely and so did the random search.

`

**TABLE 8.7.** SET OF HYPERPARAMETERS OF DIFFERENT MODELS

| Sl no | Model | Set of Hyperparameters |
|---|---|---|
| 1 | Ridge Regression | {'**alpha**': Uniform(0.1,1000)} |
| 2 | Logistic Regression (L2 regularized) | {'**C**':Uniform(0.00001,1.0)} |
| 3 | Adaboost Regression Adaboost Classification | {'**learning_rate**':Uniform(0.00001,1.0) '**max_depth**':Uniform_int(1,7) '**n_estimators**':Uniform_int(1,1000)} |
| 4 | RandomForest Regression RandomForest Classification | {'**max_features**':Uniform(0.0,1.0) '**n_estimators**':Uniform_int(1,1000)} |
| 5 | GradientBoosting Regression GradientBoosting Classification | {'**learning_rate**':Uniform(0.00001,1.0) '**max_depth**':Uniform_int(1,7) '**n_estimators**':Uniform_int(1,1000)} |
| 6 | XgBoost Regression XgBoost Classification | {'**learning_rate**':Uniform(0.00001,1.0) '**max_depth**':Uniform_int(1,7) '**n_estimators**':Uniform_int(1,1000)} |
| 7 | LightGBM Regression LightGBM Classification | {'**learning_rate**':Uniform(0.00001,1.0) '**max_depth**':Uniform_int(1,7) '**n_estimators**':Uniform_int(1,1000)} |

**TABLE 8.8.** PERFORMANCE OF MODELS BASED ON FINE-TUNED HYPERPARAMETERS USING DIFFERENT ALGORITHMS ON THE STEEL INDUSTRY DATASET
(METRIC: RMSE)

| Model | TPE | Genetic Algo. | Random |
|---|---|---|---|
| Ridge Regression | 3.895 | 8.80 | 8.85 |
| RandomForest Regression | 0.8861 | 0.865 | 0.865 |
| GradientBoosting Regression | **0.829** | 0.871 | 0.871 |
| Xgboost Regression | 1.058 | **0.778** | 0.760 |
| LightGBM Regression | 1.02 | 1.06 | **0.757** |
| Adaboost Regression | 2.77 | 2.52 | 2.52 |

`

A higher RMSE value for ridge regression clearly suggests that the dataset has a nonlinear relationship between the predictor and the target variables. Thus, this model acts as a basic standard to compare how nonlinear models are performing vis-à-vis a linear model. A significant drop in the RMSE score by nonlinear models shows the power of the model to extract the nonlinear pattern from within the dataset.

For the classification tasks, the Cohen Kappa score is used because the classes are not balanced. Since the Kappa score takes into account the chance factor in its calculation, this is more suitable than the accuracy score while evaluating performances in classification with imbalanced classes.

**TABLE 8.9.** PERFORMANCE OF MODELS BASED ON FINE-TUNED HYPERPARAMETERS USING DIFFERENT ALGORITHMS ON GAS TURBINE DATASET (METRIC: RMSE)

| Model | TPE | Genetic Algo. | Random |
|---|---|---|---|
| Ridge Regression | 1.420 | 1.421 | 1.420 |
| RandomForest Regression | 1.215 | **1.205** | 1.206 |
| GradientBoosting Regression | 1.210 | 1.232 | **1.179** |
| Xgboost Regression | **1.20** | 1.281 | 1.285 |
| LightGBM Regression | 1.243 | 1.267 | 1.245 |
| Adaboost Regression | 1.446 | 1.232 | 1.276 |

**TABLE 8.10.** PERFORMANCE FINE-TUNED MODELS USING DIFFERENT ALGORITHMS ON THE ADULT DATASET (METRIC: COHEN KAPPA)

| Model | TPE | Genetic Algo. | Random |
|---|---|---|---|
| Logistic Regression | 0.536 | 0.534 | 0.534 |
| RandomForest Regression | 0.76 | 0.758 | 0.763 |
| GradientBoosting Regression | **0.793** | 0.790 | 0.789 |
| Xgboost Regression | 0.787 | 0.790 | **0.793** |
| LightGBM Regression | 0.789 | **0.791** | 0.790 |
| Adaboost Regression | 0.788 | 0.787 | 0.789 |

`

TABLE 8.11. PERFORMANCE OF FINE-TUNED MODELS USING DIFFERENT ALGORITHMS ON BEAN DATASET (METRIC: COHEN KAPPA)

| Model | TPE | Genetic Algo. | Random |
|---|---|---|---|
| Logistic Regression | 0.944 | 0.944 | 0.945 |
| RandomForest Regression | 0.950 | 0.951 | 0.950 |
| GradientBoosting Regression | 0.953 | 0.953 | 0.952 |
| Xgboost Regression | **0.954** | **0.954** | **0.953** |
| LightGBM Regression | 0.951 | 0.951 | 0.951 |
| Adaboost Regression | 0.952 | 0.953 | **0.953** |

Thus, based on the above experimentations, it can be said that random search tends to give a very good performance even though it does not take into account the outcomes of the previous runs. However, it should be understood that the number of hyperparameters taken into account is less in boosting-based algorithms. Three hyperparameters are tuned using the algorithms. This would also mean that the search space is smaller (3 dimensional) and in smaller search space, the probability of getting an optimal point increase with the increase in the samples. If more than 3 hyperparameters are considered, (say 6 hyperparameters), the search space will increase significantly and the probability of getting an optimum or near optimum solution will reduce while working with only 100 samples. In those situations, intelligent search algorithms may dominate the random search algorithm. However, that study is not included in this work. Readers can try this out with more computational resources.

## Conclusion

The study focuses on three prominent algorithms for hyperparameter tuning. Two different tasks are considered in this study, i.e., regression and classification. Linear models as well as nonlinear models are trained to see the relative performances. The nonlinear models outperformed the linear models (ridge regression and logistic regression) by large margins (in three instances, 2 regressions and one classification task) suggesting that as the hyperparameters are

`

properly tuned, the nonlinear models map the pattern more accurately to deliver superior predictions. This tuning process is time-consuming and computationally intensive. The three algorithms used are TPE, Genetic Search, and Random Search. Quite interestingly, for regression analysis, Random Search provided the best results whereas, for the classification tasks, TPE turned out to be the best. Hence, there is no clear winner. However, TPE and Genetic search make use of the previous outcomes in an intelligent way but random search relies entirely on having the good (or best) solution based on the sample collection. Thus, as the search space is increased, the performance of random search may degrade rather quickly if the sample size is not increased.

`

`

`

`

`

`